%% file: dyn_main.tex
\documentclass[10pt,twocolumn,letterpaper]{article}

\usepackage[pagenumbers]{cvpr} 

\input{preamble}

\definecolor{cvprblue}{rgb}{0.21,0.49,0.74}
\usepackage[pagebackref,breaklinks,colorlinks,allcolors=cvprblue]{hyperref}

\def\paperID{9504} 
\def\confName{CVPR}
\def\confYear{2026}

\title{SV-GS: Sparse View 4D Reconstruction with Skeleton-Driven Gaussian Splatting}

\author{Jun-Jee Chao\\
University of Minnesota\\
{\tt\small chao0107@umn.edu}
\and
Volkan Isler\\
The University of Texas at Austin\\
{\tt\small isler@cs.utexas.edu}
}

\begin{document}
\maketitle

\begin{strip}\centering
\includegraphics[width=0.99\linewidth,trim={2cm 7.9cm 2.9cm 5.3cm},clip]{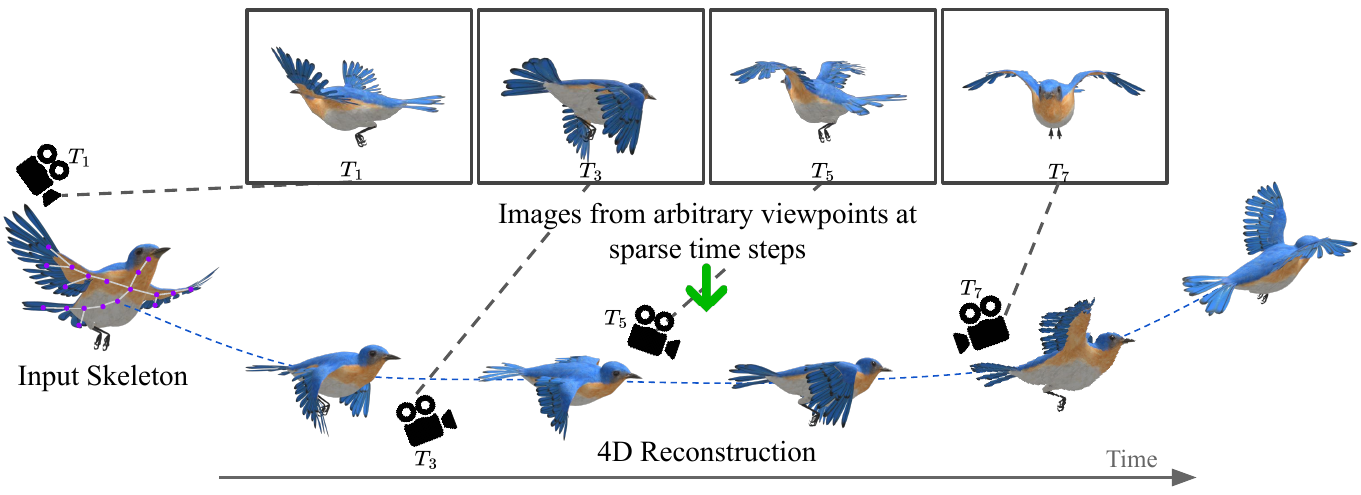}
\captionof{figure}{We study the problem of 4D reconstruction from sparse observations. Our method takes the following as input: (a)~A set of posed RGB images of an articulated target, captured at sparse time steps (up to 20x fewer than existing methods) from arbitrary viewpoints; (b)~An annotated skeleton graph only at the first frame; (c)~An initial static 3D reconstruction, derived either from multi-view images or a pre-trained image-to-3D diffusion model. Our goal is to produce a continuous 4D reconstruction of the dynamic target. }
\label{dyn_img:teaser}
\end{strip}



\input{dyn_sec/abstract}    
\input{dyn_sec/intro}

\input{dyn_sec/related_works}
\input{dyn_sec/method}

\input{dyn_sec/exp}
\input{dyn_sec/conclusion}


\paragraph{\fontsize{11.5}{12.5}\selectfont{Acknowledgment}.}
This material is based upon work supported in part by the National Science Foundation under Grant No. 2504906 and National Research Foundation of Korea (NRF) grant (MSIT) No. RS-2024-00462874.

{
    \small
    \bibliographystyle{ieeenat_fullname}
    \bibliography{main}
}


\end{document}

%% file: preamble.tex
\newcommand{\mymat}[1]{\mathbf{#1}}
\newcommand{\myset}[1]{\mathcal{#1}}

\newcommand{\eref}[1]{Equation (\ref{#1})}
\newcommand{\secref}[1]{Section~\ref{#1}}
\newcommand{\tabref}[1]{Table~\ref{#1}}

\newcommand{\figref}[1]{Fig.~\ref{#1}}

\newcommand{\ourmethod}{\textbf{SV-GS}}

\usepackage[accsupp]{axessibility}  
\usepackage{times}
\usepackage{comment}
\usepackage{epsfig}
\usepackage{graphicx}
\usepackage{amsmath,amssymb,amsfonts,amsthm,bm}
\usepackage{xcolor}
\usepackage{multicol}
\usepackage{multirow}
\usepackage{booktabs}
\usepackage{array}
\usepackage{algorithm}
\usepackage[noend]{algpseudocode}
\usepackage{makecell}
\usepackage{paralist}
\usepackage{soul}
\usepackage{wrapfig}
\usepackage{epsfig}
\usepackage{graphicx}
\usepackage{adjustbox}
\usepackage{cuted}
\usepackage{capt-of}
\usepackage{tikz} 
\usepackage{enumitem}
\usepackage{float}
\usepackage{capt-of,etoolbox}
\usepackage[skip=2pt]{caption}
\usepackage{placeins}

%% file: dyn_sec/abstract.tex
\begin{abstract}
Reconstructing a dynamic target moving over a large area is challenging. Standard approaches for dynamic object reconstruction require dense coverage in both the viewing space and the temporal dimension, typically relying on multi-view videos captured at each time step.
However, such setups are only possible in constrained environments.
In real-world scenarios, observations are often sparse over time and captured sparsely from diverse viewpoints (e.g., from security cameras), making dynamic reconstruction highly ill-posed. 
We present \ourmethod, a framework that simultaneously estimates a deformation model and the object’s motion over time under sparse observations.
To initialize \ourmethod, we
leverage a rough skeleton graph and an initial static reconstruction as inputs to guide motion estimation. (Later, we show that this input requirement can be relaxed.)
Our method optimizes a skeleton-driven deformation field composed of a coarse skeleton joint pose estimator and a module for fine-grained deformations.
By making only the joint pose estimator time-dependent, our model enables smooth motion interpolation while preserving learned geometric details. 
Experiments on synthetic datasets show that our method outperforms existing approaches under sparse observations by up to 34\% in PSNR, and achieves comparable performance to dense monocular video methods on real-world datasets despite using significantly fewer frames. Moreover, we demonstrate that the input initial static reconstruction can be replaced by a diffusion-based generative prior, making our method more practical for real-world scenarios.

\end{abstract}

%% file: dyn_sec/intro.tex
\section{Introduction}
\label{dyn_sec:intro}

Reconstructing dynamic targets from images is a long-standing computer vision problem, with applications in motion analysis~\cite{liu2023building,chao2025part}, AR/VR~\cite{yan2025instant}, and dynamic scene understanding~\cite{wang2023root}. While recent progress in neural~\cite{pumarola2021d,cao2023hexplane,uzolas2023template} and Gaussian-based representations~\cite{Wu_2024_CVPR,wan2024template,huang2024sc,yao2025riggs} have shown impressive results, most methods rely on monocular or multi-view videos with dense temporal coverage, where rich motion cues and correspondences are available. 

In real-world scenarios, however, such dense observations are not always accessible. For example, surveillance cameras often capture moving objects sparsely over time, especially in cluttered environments.
Moreover, when multiple cameras are available, their viewpoints can differ drastically, and the observed targets may exhibit significant motion and self-occlusion between observations.
Under this setting, temporal correspondences are difficult to establish, as appearance can change dramatically across sparse observations, making dynamic reconstruction highly ill-posed.

In this paper, we address this challenging setting of articulated dynamic reconstruction from sparse temporal observations, where only a few posed images from arbitrary viewpoints are available as illustrated in \figref{dyn_img:teaser}.
To solve this highly ill-posed problem, we consider a setting where we have access to additional structural information. 
Initially, we assume that a rough skeleton graph and a static reconstruction at the first frame are available. 
The initial reconstruction can be can be obtained from a standard multi-view setup ~\cite{schoenberger2016sfm, schoenberger2016mvs, kerbl3Dgaussians}.
Later on in \secref{dyn_sec:diffusion}, we will show how this assumption can be relaxed with a pre-trained generative model~\cite{liu2023zero, shi2023mvdream, tang2023dreamgaussian} using only a single image. 
Despite this additional information, the task remains difficult as the inputs do not yield a complete rigged model---the skeleton annotation can be noisy and contains only node positions and connectivity, while the joint poses, skinning weights, and point-to-part associations remain unknown. 

We present \ourmethod{} which, given the input skeleton graph and initial static reconstruction, learns a skeleton-driven deformation field that models coherent motion under sparse supervision. Our deformation field consists of a coarse skeleton joint pose estimator and a module that models fine-grained motion deformations. By allowing only the joint pose estimator to be time-dependent, our model enables smooth test-time motion interpolation while preserving learned local deformation details.
Experiments demonstrate that state-of-the-art (SOTA) dynamic reconstruction methods degrade significantly in this sparse setting, while \ourmethod{} achieves better reconstruction quality. 
Furthermore, we show that the need for multi-view initialization can be relaxed using a diffusion-based generative prior, enabling dynamic reconstruction in real-world scenarios. 
Our contributions can be summarized as follows. 
\begin{compactitem} 
\item  We perform articulated dynamic reconstruction from sparse temporal observations, where only a few frames from arbitrary viewpoints are available.
\item We present a skeleton-driven deformation field that enables smooth motion interpolation under sparse supervision, and demonstrate that a pre-trained diffusion prior can be incorporated to fill in missing information.
\item Experiments show that our method outperforms SOTA methods by up to 34\% in PSNR on synthetic datasets with sparse observations, and achieves comparable performance to dense monocular video methods on real-world datasets with significantly fewer frames.
\end{compactitem}

%% file: dyn_sec/related_works.tex
\section{Related Work}
\label{dyn_sec:rel_work}

\begin{figure}[b]
    \centering
    \includegraphics[width=\columnwidth,trim={5.2cm 5.5cm 6.2cm 5.7cm},clip]{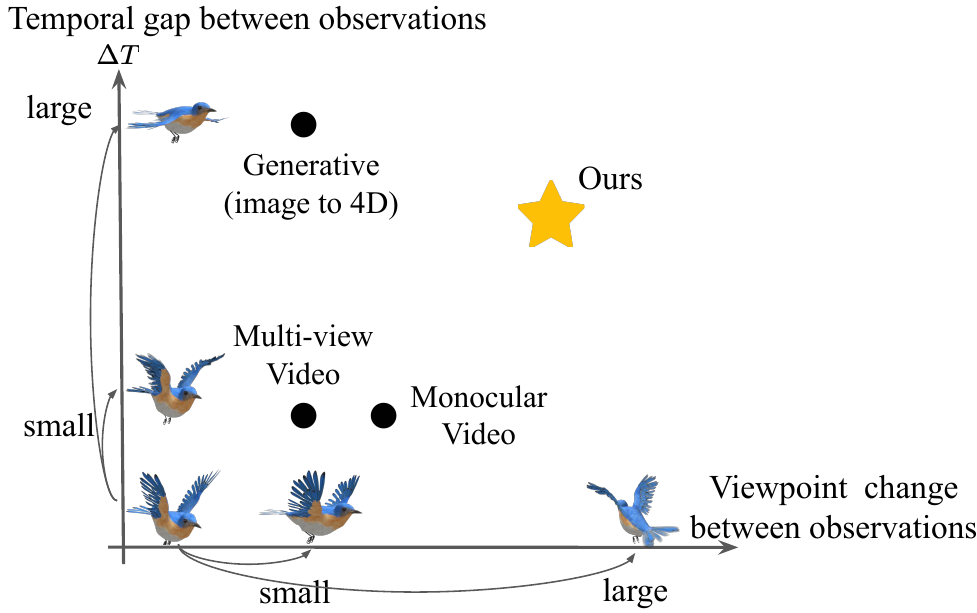}
    \caption{Comparison of input configurations across dynamic reconstruction methods. Multi-view and monocular video methods assume small viewpoint changes and dense temporal observations, whereas our method handles sparse temporal observations with large viewpoint variations. Generative methods attempt to synthesize the full motion from a static state.}
    \label{dyn_img:related_works}
\end{figure}

We review related works on dynamic scene reconstruction and articulated object modeling.
As most existing methods rely on video inputs (see \figref{dyn_img:related_works}), we also discuss recent generative approaches that are related to our sparse-view setting. We further quantify the difficulty of our setup using the metric from~\cite{gao2022monocular} in the supplementary material.

\textbf{Dynamic scene modeling.}
Some earlier methods apply explicit mesh representation~\cite{dou2016fusion4d,broxton2020immersive} or implicit neural volumes~\cite{lombardi2019neural} to model dynamic scenes from multi-view videos, leveraging the dense spatial and temporal information. After NeRF~\cite{mildenhall2021nerf} was introduced, the field of novel view synthesis became even more popular. D-NeRF~\cite{pumarola2021d} and many concurrent works extend the static NeRF representation to dynamic scene by optimizing an additional time-dependent deformation field~\cite{li2021neural,park2021nerfies,park2021hypernerf,tretschk2021non, guo2023forward, xian2021space}, or by directly modeling the 4D space~\cite{fridovich2023k,cao2023hexplane, shao2023tensor4d}. 

3D Gaussian Splatting (3DGS)~\cite{kerbl3Dgaussians} is another scene representation that has gained popularity due to its fast rendering speed. Many recent works adapt 3DGS for dynamic scene reconstruction~\cite{luiten2024dynamic,Wu_2024_CVPR,wan2024superpoint,huang2024sc,liu2025modgs,yang2024deformable,kerbl3Dgaussians, kratimenos2024dynmf}. 
4DGS~\cite{Wu_2024_CVPR} decouples the scene into a static 3DGS and a deformation field represented with multi-resolution hex-planes~\cite{cao2023hexplane}. Recently, a line of work attempts to model the dynamic scene with a more controllable representation by using a sparse set of parameters to represent the dense deformation~\cite{wan2024superpoint, huang2024sc}.
However, most existing methods rely on monocular videos with dense temporal information, which is unavailable in our sparse observation setup (\figref{dyn_img:related_works}). Moreover, without structural constraints, these approaches can produce noisy deformations that fail to preserve the object’s structure under sparse supervision.

\par
\vspace{6pt}
\noindent\textbf{Articulated object reconstruction.}
To model dynamic articulated objects, some methods leverage category-specific priors. For example, SMPL~\cite{SMPL:2015} focuses on human body modeling, and MANO~\cite{romero2017embodied} focuses on human hands. Another line of work tackles the animal category where a kinematic structure is shared among different instances~\cite{zuffi20173d, wu2022casa, yao2022lassie, NEURIPS2023_a99f50fb, wu2023magicpony, lei2024gart}. However, many of these works focus on part discovery from a single image instead of reconstructing the continuous motion for novel view synthesis~\cite{NEURIPS2023_a99f50fb, yao2022lassie,wu2023magicpony}. 


More general category-agnostic methods have been explored~\cite{noguchi2022watch, yang2022banmo, zhang2024bags, wan2024template, yao2025riggs}. Many of these methods focus on simultaneously modeling the dynamic target and extracting the underlying kinematic structure from video input. 
SK-GS~\cite{wan2024template} extends SP-GS~\cite{wan2024superpoint} by first grouping the 3DGS with similar motion into superpoints. Then, they extract a skeleton model from the superpoints based on relative motion and proximity.
Similarly, built upon SC-GS~\cite{huang2024sc}, RigGS~\cite{yao2025riggs} first estimates a set of sparse control points to model the dynamic scene, then the kinematic skeleton is estimated from the motion of these control points. While these methods learn skeleton-driven deformation for 3DGS which is similar to our setup, they take continuous monocular videos as input, and do not perform well when only sparse images are available. Moreover, we show in \secref{dyn_sec:exp_main} that with the same initialization and skeleton input, these methods designed for monocular video fail when only sparse images are available.

\par
\vspace{6pt}
\noindent\textbf{Scene reconstruction with generative priors.}
A pre-trained generative model can potentially be applied to fill in the missing information from sparse observations.
Many recent works apply pre-trained diffusion models for static scene reconstruction from one or more images~\cite{liu2023zero, liu2024one, tang2023dreamgaussian, shi2023mvdream}. To extend from static to dynamic scene, a popular approach is to apply the SDS loss~\cite{pooledreamfusion} to guide the motion with a pre-trained video diffusion model~\cite{zhao2023animate124, li2025articulated, xu2024comp4d,bahmani20244d,ling2024align,zeng2024stag4d,zheng2024unified}. However, these methods focus on the generative setup where the generated motion is expected to be smooth and reasonable but does not need to match any ground truth. On the contrary, our problem setup requires us to estimate the ground truth motion from sparse observations. 


%% file: dyn_sec/method.tex
\begin{figure*}[t!]
    \centering
    \includegraphics[width=2\columnwidth,trim={1.0cm 7.5cm 1.0cm 5.35cm},clip]{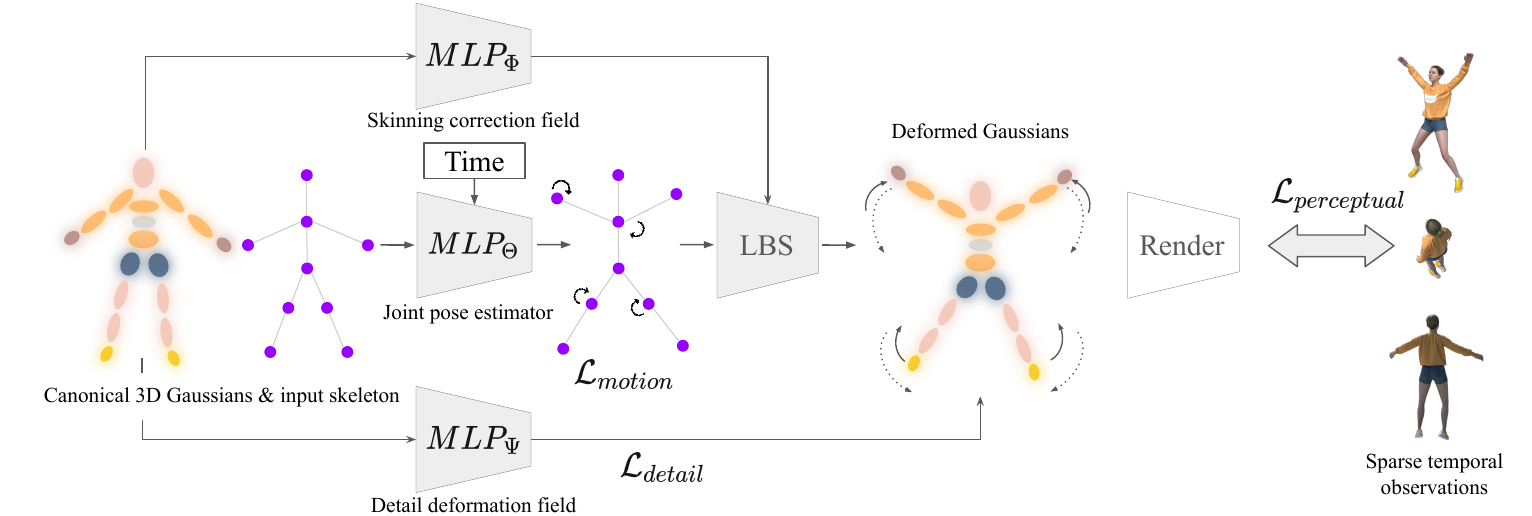}
    \caption{Given canonical 3D Gaussians and an input skeleton, \ourmethod{}  first predicts time-dependent joint poses, regularized with $\mathcal{L}_{motion}$ for temporal smoothness. With the predicted skeleton poses, the canonical Gaussians are then transformed via Linear Blend Skinning using learnable per-bone radii and a skinning correction field. Finally, a detail deformation field refines the transformed Gaussians. All parameters are optimized by minimizing the perceptual loss between the rendered and observed images. }
    \label{dyn_img:method}
    \vspace{-4pt}
\end{figure*}

\section{Method}
\label{dyn_sec:method}

Our goal is to reconstruct an articulated dynamic target from sparse temporal observations $\myset{I} = \{I_t\}_{t\in[0,1]}$, where each time step consists of only a single posed image captured from an arbitrary viewpoint. We present \ourmethod{}, which assumes access to a skeleton structure $\myset{F}$ as input.  The skeleton specifies the 3D locations of $J$ nodes and their parent–child connectivity, which can be obtained through human annotation or estimated using an off-the-shelf method~\cite{liu2025riganything, RigNet}.
As illustrated in \figref{dyn_img:method}, \ourmethod{} starts from building an initial static 3D reconstruction of the target. Then we learn a skeleton-driven deformation field that models continuous articulation and motion over time, under sparse temporal supervision. 

\subsection{Scene representation}

\par
\noindent\textbf{Initial Static 3D Gaussians.} We adopt 3D Gaussian Splatting (3DGS)~\cite{kerbl3Dgaussians} as our scene representation for its fast optimization speed and explicit, physically interpretable parameterization. 3DGS represents a scene with a collection of Gaussian primitives $\myset{G} = \{ g_i \}_{i \in 1,...,N}$, where each Gaussian $g_i$ is defined by a center $\mu_i$, a rotation matrix represented with quaternion $q_i$, a scaling vector $s_i$, an opacity value $\sigma_i$, and a set of spherical harmonics coefficients $sh_i$ determining the view-dependent color. Given a camera pose, we can render an image from $\myset{G}$, where the pixel color is determined by $\alpha$-blending along the ray direction:

\begin{equation}
color = \sum_k c_k \alpha_k \prod_{j=1}^{k-1} (1 - \alpha_j)
\end{equation}
where k is the index of the Gaussians sorted by depth along the viewing direction, and $c_k$ is the view-dependent color evaluated from the spherical harmonics coefficients.
The $\alpha$ value is the opacity $\sigma$ weighted by the projected 2D Gaussian distribution from the 3D space onto the 2D plane. 

In this paper, we assume the initial static 3DGS can be obtained  either from multi-view images or potentially from a pre-trained image-to-3D diffusion model. In the multi-view setup, we follow the standard pipeline~\cite{kerbl3Dgaussians} to optimize the Gaussian parameters by minimizing the perceptual loss between the rendered images and the ground truth images. 
We further showcase in \secref{dyn_sec:diffusion} that the multi-view initialization can potentially be replaced with a pre-trained generative model. More details can be found in \secref{dyn_sec:diffusion} and the supplementary material.

\par 
\vspace{6pt}
\noindent\textbf{Skeleton-Driven Deformation.}
Given the initial static 3DGS $\myset{G}$ and an annotated skeleton graph $\myset{F}$, our goal is to learn a deformation field that transforms the initial $\myset{G}$ to match the observed images at the corresponding sparse time steps. Furthermore, the learned deformation enables continuous motion synthesis for intermediate time steps without direct observations.
Note that the input skeleton graph can be noisy and contains only the 3D positions of the nodes and their connectivity, without point-to-part associations or joint parameters.

To derive a deformation that is constrained by the input skeleton  while also remaining flexible to match the sparse observations, we draw inspiration from learnable Linear Blend Skinning (LBS) techniques~\cite{magnenat1989joint,yang2022banmo,yao2025riggs,li2025articulated}. 
Specifically, we adopt an MLP to model the time-dependent local rotation $q^t_j $ (represented using quaternions) for each joint $j$ in the skeleton, along with a local translation $p^t \in \mathbb{R}^3$ only for the root joint. 

\begin{equation}
q^t, p^t = MLP_{\Theta}(\gamma(t))
\end{equation}
where $\gamma(\cdot)$ denotes the positional encoding~\cite{mildenhall2021nerf}. 
The local rotations are defined for each joint in the local frame, therefore, given the parent–child hierarchy in the skeleton graph $\myset{F}$, we compute the global transformation of each joint using forward kinematics~\cite{denavit1955kinematic}

\begin{equation}
\hat{\mymat{R}^t}, \hat{T^t} = fk(\myset{F}, q^t, p^t)
\end{equation}
where $\hat{\mymat{R}^t_j}$ and $\hat{T^t_j}$ denote the global rotation (represented as $3\times3$ matrix) and translation of joint $j$ at time $t$ respectively. $fk(\cdot)$ is the forward kinematics operation that propagates the local transformation of each joint to all child joints. 

Next, to guide the Gaussian primitives with the estimated joint poses, we derive a fine-grained motion field based on a learnable LBS deformation. We first construct $B$ bones, where each bone $b_j$ corresponds to the edge connecting joint $j$ and it's parent~\cite{yao2025riggs,uzolas2023template}. 
Each Gaussian center $\mu_i$ in the canonical static state is transformed to time $t$ as:
\begin{equation}
\mu_i^t = \sum_{j=1}^B w_{i,j}(\hat{\mymat{R}^t_j} \mu_i + \hat{T^t_j})
\end{equation}
where $w_{i,j}$ is the learnable skinning weight satisfying $\Sigma_j w_{i,j} = 1$. The rotation part of the Gaussian primitive is similarly approximated by the weighted sum: $\sum_{j=1}^B w_{i,j}\hat{\mymat{R}^t_j}\mymat{R}_i$. 

\par 
\vspace{6pt}
\noindent\textbf{Learnable Skinning Weights.}
Since the input skeleton can be noisy and lacks skinning and deformation information, we model the skinning effect of each bone as a Radial Basis Function (RBF) kernel in the canonical (static) state. Moreover, to account for the noise in the input skeleton, we learn a position-dependent correction field $MLP_{\Phi}$ also in the canonical state. 
Formally, we compute normalized weights as:
\begin{equation}
w_{i,j} = \frac{\hat{w_{i,j}}}{ \sum_{j=1}^B \hat{w_{i,j}}},
\end{equation}
where
\begin{equation}
\hat{w_{i,j}} = \Delta w_{i,j} \ exp\left(- \frac{d_{i,j}^2}{2r_j^2}\right)
\end{equation}
Here $d_{i,j}$ denotes the the distance between the Gaussian center $\mu_i$ and bone $b_j$ in the canonical frame, and $r_j$ is the learnable influence radius for each bone $j$. Moreover, the correction field $\Delta w_{i,j}$ is parameterized with a MLP:

\begin{equation}
\Delta w_{i,j} = MLP_{\Phi}(\gamma(\mu_i))
\end{equation}
where $\gamma(\cdot)$ again denotes the positional encoding~\cite{mildenhall2021nerf} for the Gaussian center $\mu_i$.

\par 
\vspace{6pt}
\noindent\textbf{Detail Deformation.}
The above skeleton-driven deformation captures coarse articulated motion by propagating the joint transformations to the Gaussian primitives. However, the skeleton is sparse by nature and cannot account for fine-grained non-rigid deformations. Inspired by~\cite{yao2025riggs}, we include an additional pose-dependent detail deformation field $MLP_\Psi$ to refine the local details. For each Gaussian, we predict a small offset by considering the Gaussian center in the canonical frame and the predicted joint poses at that time step. Therefore, the final Gaussian center at time $t$ is:

\begin{equation}
\hat{\mu_i^t } = \mu_i^t  + MLP_\Psi(\gamma(\mu_i), \mymat{R}^t)
\end{equation}

\begin{figure*}[!htb]
    \centering
    \includegraphics[width=2\columnwidth,trim={2.6cm 5.8cm 2.9cm 4.75cm},clip]{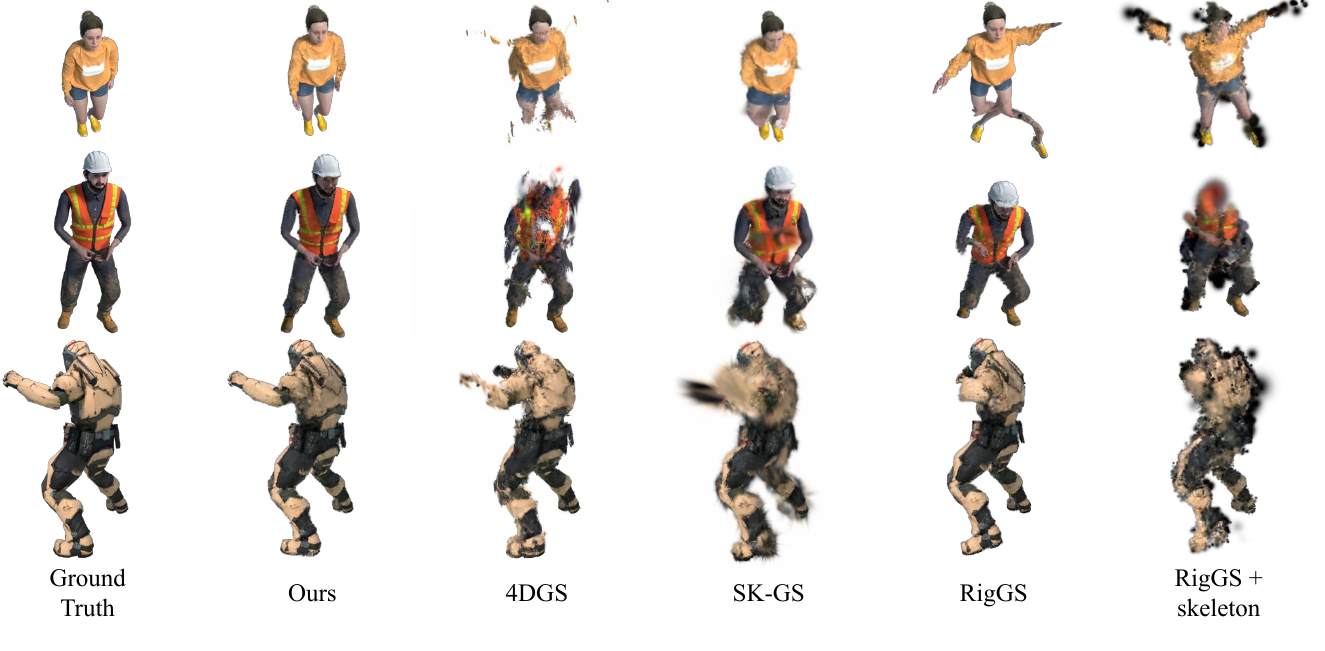}
    \caption{Qualitative results on the D-NeRF dataset~\cite{pumarola2021d} downsampled at 0.1 intervals, yielding 11 frames per motion sequence (up to $20\times$ fewer than the original). 
    We compare our method with SOTA methods including 4DGS~\cite{Wu_2024_CVPR}, SK-GS~\cite{wan2024template}, and RigGS~\cite{yao2025riggs}. Additionally, we modify RigGS~\cite{yao2025riggs} to take in the same skeleton input as ours. 
    Despite all methods being initialized with the same multi-view images at $t=0$, existing methods produce noisy deformations and fail to preserve object structure given only sparse temporal observations. }
    \label{dyn_img:dnerf}
    \vspace{-4pt}
\end{figure*}

\subsection{Optimization}
\label{dyn_sec:optimization}

The trainable parameters of our deformation field include the joint local pose predictor $MLP_\Theta$, the bone influence radii $r_j$, the skinning correction field $MLP_\Phi$, and the detail deformation field $MLP_\Psi$. During training the deformation parameters, we keep the parameters of the static canonical Gaussians $\myset{G}$ fixed. All deformation parameters are jointly optimized by minimizing the following loss:

\begin{equation} \label{dyn_eq:loss_func}
\mathcal{L} = \lambda_{1}\mathcal{L}_{perceptual} + \lambda_{2}\mathcal{L}_{motion} + \lambda_{3}\mathcal{L}_{detail}
\end{equation}

The main objective is to enforce the deformed Gaussians to match the observed images when rendered from the corresponding viewpoints. We follow the perceptual loss used in 3DGS~\cite{kerbl3Dgaussians}, where $\mathcal{L}_{perceptual}$ is a combination of $\mathcal{L}_{1}$ loss and D-SSIM loss. 

However, since only one image observation is available at each sparse time step, regions without direct supervision may undergo unstable or noisy deformation. To address this, we introduce two regularization terms that constrain the skeleton motion and the detail deformation field.

\par 
\vspace{6pt}
\noindent\textbf{Motion Regularization.} 
Since the joint poses are defined in their respective local frames, we can directly enforce temporal smoothness by minimizing the Laplacian of the predicted values with respect to time:

\begin{equation}
\mathcal{L}_{motion} = \frac{1}{T J}\sum_t^T \sum_j^J \left| q_j^{t-1} - 2q_j^t + q_j^{t+1} \right |
\end{equation}
where $T$ is uniformly sampled between $[0,1]$. 
This regularization helps mitigate the ambiguity caused by self-occlusions under single-view supervision at each time step, preventing $MLP_\Theta$ from producing abrupt pose changes and encouraging temporally coherent motion.

\par 
\vspace{6pt}
\noindent\textbf{Detail Deformation Regularization.} The detail deformation field $MLP_\Psi$ is defined in the canonical frame to model small offsets for each Gaussian primitive such that the rendered images reflect finer motion details. Since this field is not intended to cause large displacements, we apply an $\mathcal{L}_2$ regularization term on the predicted offsets:

\begin{equation}
\mathcal{L}_{detail} = \frac{1}{N}\sum_i^N \left\| MLP_\Psi(\gamma(\mu_i), \mymat{R}^t) \right\|^2_2
\end{equation}

\subsection{Inference}

Our ultimate goal is to reconstruct a continuous motion sequence from sparse observations. Since the model is only supervised at a few discrete time steps, the learned $MLP$ may produce temporally inconsistent or jittery motions when queried at unseen time steps. To mitigate this issue, we design the deformation field such that only the local pose prediction $MLP_\Theta$ depends explicitly on time. This allows us to effectively perform interpolation for the joint poses at unseen intermediate time steps while preserving the effect of the skinning correction field and detail deformation field. We show in the supplementary video that our method generates smooth and coherent motion even under sparse temporal supervision.

%% file: dyn_sec/exp.tex
\begin{figure}[t!]
    \centering
    \includegraphics[width=\columnwidth,trim={1.2cm 8.2cm 1.2cm 4.2cm},clip]{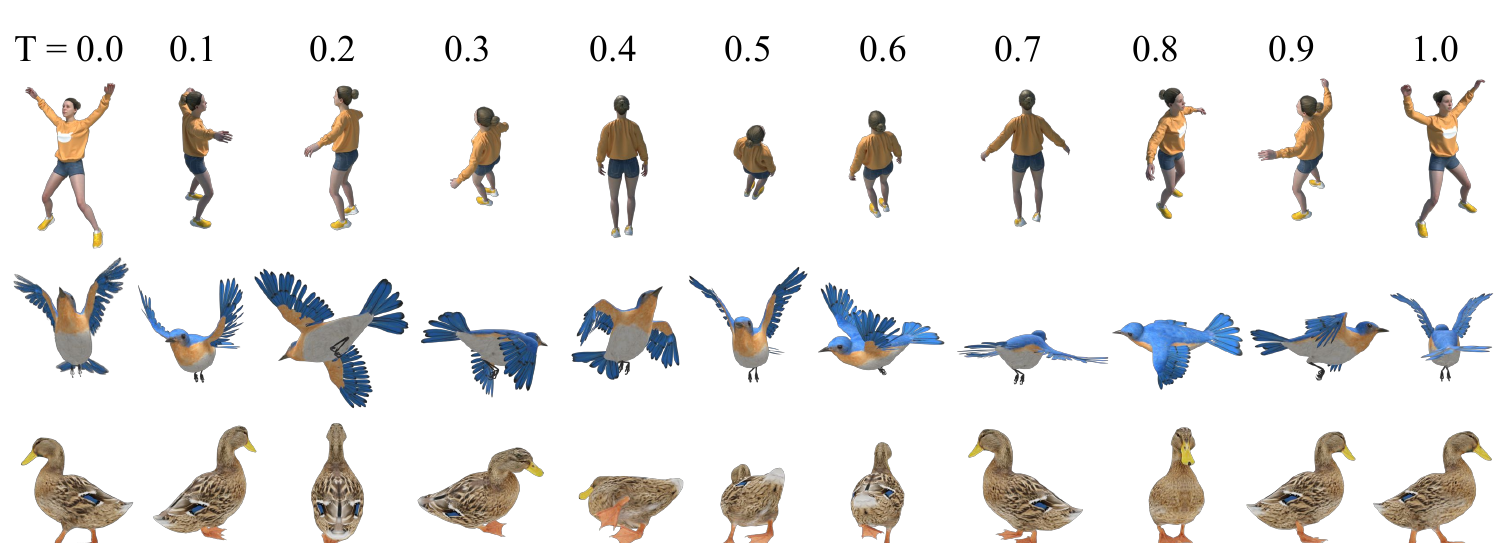}
    \caption{We show \textbf{all} input views from the downsampled dataset (up to $20\times$ fewer frames than the original), illustrating the challenges of establishing correspondences under sparse observations, large viewpoint changes, and self-occlusions.}
    \label{dyn_sec:inputs}
    \vspace{-4pt}
\end{figure}

\section{Experiments}
\label{dyn_sec:exp}

We first compare our method against existing approaches on novel view synthesis under sparse temporal observations, given a multi-view reconstruction at the initial state (\secref{dyn_sec:exp_main}). 
We then demonstrate that the multi-view initialization can be replaced by a pre-trained diffusion-based generative model (\secref{dyn_sec:diffusion}), highlighting the potential of our approach in more challenging scenarios.

\subsection{Experimental Setup}
\label{dyn_sec:exp_setup}
\noindent\textbf{Datasets.} 
Our experiments are mainly conducted on three datasets: D-NeRF~\cite{pumarola2021d}, DG-Mesh~\cite{liudynamic}, and ZJU-MoCap~\cite{peng2021neural}. D-NeRF~\cite{pumarola2021d} contains 6 synthetic scenes after excluding those with multiple objects or inconsistent motion between training and testing~\cite{huang2024sc}. DG-Mesh~\cite{liudynamic} includes 5 synthetic sequences of articulated animal models. 
We normalize the time steps to the $[0,1]$ range and uniformly subsample frames at $0.1$ intervals, resulting in $11$ image observations per sequence, where each observation is captured from an arbitrary camera viewpoint at that time step as illustrated in \figref{dyn_sec:inputs}. 
This corresponds to up to $20\times$ fewer time frames compared to the original datasets.
Following~\cite{yao2025riggs}, we evaluate our approach on 6 real-world sequences from the ZJU-MoCap dataset~\cite{peng2021neural}. 
Since ZJU-MoCap contains longer motion sequences with more complex movements, we downsample the frame rate to $1/10$ from the original, where each time step is again arbitrarily selected from the training views.
To further demonstrate the generalization ability of our method on in-the-wild data, we additionally test on the camel scene from the DAVIS dataset~\cite{Perazzi_CVPR_2016}, where no camera pose information is provided.

\par 
\vspace{6pt}
\noindent\textbf{Metrics.} 
We evaluate the quality of novel view synthesis using three standard metrics: Peak Signal-to-Noise Ratio (PSNR), Structural Similarity Index (SSIM)~\cite{wang2004image}, and Learned Perceptual Image Patch Similarity (LPIPS)~\cite{zhang2018unreasonable}.

\par 
\vspace{6pt}
\noindent\textbf{Implementation Details.} The experiments are conducted on a single NVIDIA RTX 4080 GPU. Optimizations are done with PyTorch~\cite{Ansel_PyTorch_2_Faster_2024} and the ADAM optimizer~\cite{kingma2015adam}. We set $\lambda_1 = 2, \lambda_2 = 1, \lambda_3 = 1$. We run the deformation field optimization for $40{,}000$ steps for each scene, and the skeleton graph is initialized with the estimates from \cite{yao2025riggs}.
More details can be found in the supplementary material.

\subsection{Comparison with Existing Methods}
\label{dyn_sec:exp_main}

\begin{table}[t!]
\vspace{-5pt}
\centering
\caption{Quantitative results on the D-NeRF dataset~\cite{pumarola2021d} downsampled at 0.1 intervals, yielding 11 frames per motion sequence. We report the average metrics across all test cases / the mean over the worst-performing test case of each scene. $^\dagger$ indicates method initialized with the same skeleton input as ours.}
\resizebox{\columnwidth}{!}{
\begin{tabular}{cccc}
\hline
Method & SSIM $\uparrow$ & PSNR $\uparrow$  & LPIPS ${(\times100)} \downarrow$    \\ \hline
4DGS~\cite{Wu_2024_CVPR}   & 0.925 / 0.829 & 21.70 / 17.01 & 7.85 / 12.02  \\
SK-GS~\cite{wan2024template}  & 0.921 / 0.790 & 19.43 / 15.45 & 8.8 / 16.38   \\
RigGS~\cite{yao2025riggs}   & 0.897 / 0.771 & 24.23 / 19.33 & 8.28 / 13.32  \\
RigGS~\cite{yao2025riggs}$^\dagger$   & 0.839 / 0.739 & 22.63 / 19.29 & 13.82 / 18.59 \\ \hline
Ours   & \textbf{0.950} / \textbf{0.893} & \textbf{27.75} / \textbf{23.48} & \textbf{5.79} / \textbf{9.43}   \\ \hline
\end{tabular}
}
\label{dyn_table:dnerf_0.1}
\end{table}

\noindent\textbf{Synthetic Datasets.} 
Most existing dynamic scene reconstruction methods rely on either monocular video or multi-view video inputs. Therefore, to ensure a fair comparison under our sparse temporal observation setting, we provide all methods with the same multi-view posed images only at the initial time step.
We compare our method with 4DGS~\cite{Wu_2024_CVPR}, SK-GS~\cite{wan2024template}, and RigGS~\cite{yao2025riggs} for the task of novel view synthesis. 4DGS~\cite{Wu_2024_CVPR} learns a deformation field for the canonical 3DGS without any explicit structural constraint. In contrast, both SK-GS~\cite{wan2024template} and RigGS~\cite{yao2025riggs} jointly reconstruct the dynamic target and its underlying kinematic structure. Since our method takes a skeleton graph as input, we also modify RigGS~\cite{yao2025riggs} to initialize from the same skeleton for direct comparisons. 

\begin{figure}[t!]
    \centering
    \includegraphics[width=\columnwidth,trim={4.0cm 5.3cm 4.7cm 4cm},clip]{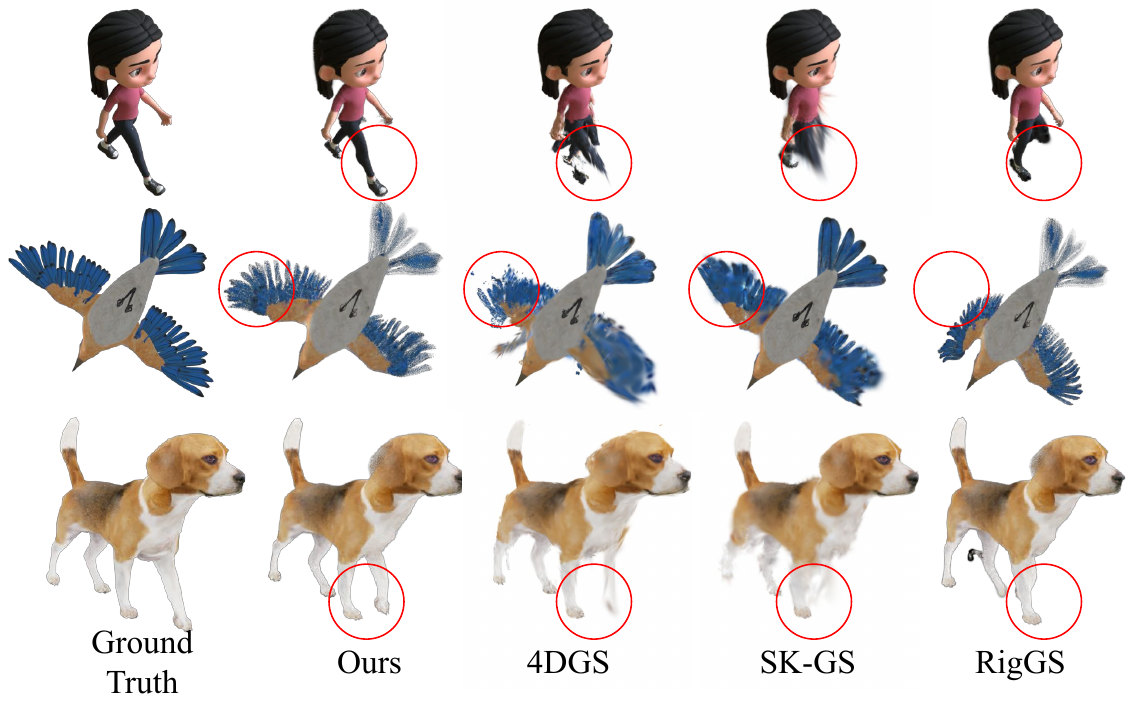}
    \caption{Qualitative result on the DG-Mesh dataset~\cite{liudynamic} downsampled at 0.05 intervals, yielding 21 frames per motion sequence. While all methods perform similarly for parts with small motion, our approach better preserves object structure and captures fine-grained motion more faithfully.}
    \label{dyn_img:dgmesh}
    \vspace{-3pt}
\end{figure}

We present qualitative results on the D-NeRF~\cite{pumarola2021d} and DG-Mesh~\cite{liudynamic} datasets in \figref{dyn_img:dnerf} and \figref{dyn_img:dgmesh}, respectively.
As shown, all baselines struggle when only sparse temporal observations are available. Without structural constraints, 4DGS~\cite{Wu_2024_CVPR} produces diverging deformations that do not preserve object structure. While SK-GS~\cite{wan2024template} and RigGS~\cite{yao2025riggs} consider skeleton constraints, they can generate inaccurate motion or skinning weights which result in blurry renderings. 
Additionally, we initialize RigGS with the same skeleton as ours, however, without careful design, the noisy skeleton and the absence of ground-truth skinning weights can lead to unstable deformations and degraded reconstruction quality. 
Quantitative results in \tabref{dyn_table:dnerf_0.1} and \tabref{dyn_table:dgmesh} confirm that our method outperforms all baselines across all evaluation metrics. 
For the DG-Mesh dataset, we evaluate two temporal downsampling configurations with intervals of $0.05$ and $0.1$, corresponding to $21$ and $11$ observable time steps respectively.
As shown in \tabref{dyn_table:dgmesh}, when more temporal observations are available, the baselines can achieve a closer SSIM score to ours, whereas our method remains robust even under severely sparse temporal inputs.

\begin{table}[t!]
\centering
\caption{Results on the DG-Mesh dataset~\cite{liudynamic} downsampled at $0.05$ and $0.1$ intervals. We present the average across all test cases / the mean over the worst-performing test case of each scene.}
\resizebox{1.0\columnwidth}{!}{
\begin{tabular}{cccc}
\hline
\multicolumn{4}{c}{DG-Mesh 0.05}      \\ \hline
Method & SSIM $\uparrow$ & PSNR $\uparrow$ & LPIPS ${(\times100)} \downarrow$ \\ \hline
4DGS~\cite{Wu_2024_CVPR}   &  0.918 / 0.822  & 23.40 / 17.68   & 7.26 / 13.23   \\
SK-GS~\cite{wan2024template}  &  0.920 / \textbf{0.833}  & 23.32 / 17.68 & 7.69 / 13.26  \\
RigGS~\cite{yao2025riggs}  &  0.879 / 0.712  & 22.81 / 16.87  & 8.36 / 15.88  \\ \hline
Ours   & \textbf{0.929} / 0.824 & \textbf{25.81} / \textbf{19.11} & \textbf{6.38} / \textbf{12.43}   \\ \hline
\hline
\multicolumn{4}{c}{DG-Mesh 0.1}       \\ \hline
Method & SSIM $\uparrow$ & PSNR $\uparrow$ & LPIPS ${(\times100)} \downarrow$ \\ \hline
4DGS~\cite{Wu_2024_CVPR}   & 0.887 / 0.774  & 21.28 / 16.07   & 8.72 / 15.41  \\
SK-GS~\cite{wan2024template} &  0.875 / 0.776 & 20.56 / 15.92  & 10.37 / 17.22 \\
RigGS~\cite{yao2025riggs} & 0.855 / 0.694 & 21.80 / 16.51  & 9.27 / 16.29   \\ \hline
Ours   & \textbf{0.900} / \textbf{0.786} & \textbf{23.76} / \textbf{17.86} & \textbf{7.59} / \textbf{13.78} \\ \hline
\end{tabular}
}
\label{dyn_table:dgmesh}
\vspace{-8pt}
\end{table}

\par
\begin{figure}[t!]
    \centering
    \includegraphics[width=\columnwidth,trim={3.6cm 9cm 4.7cm 4cm},clip]{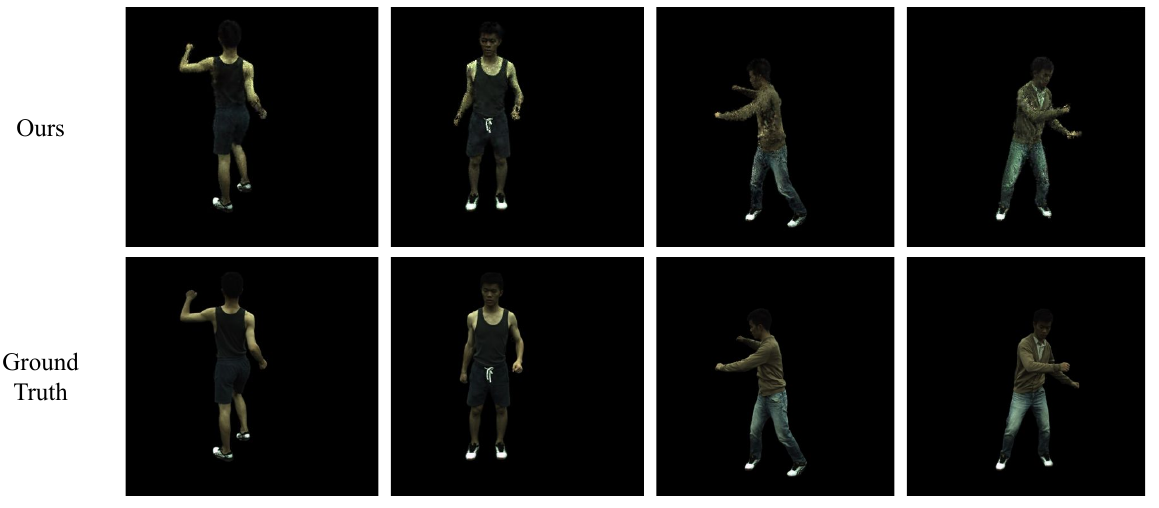}
    \caption{Qualitative result on the real-world ZJU-MoCap dataset. We use only $1/10$ of the original video frames, where each frame is sampled from an arbitrary training viewpoint at that time step.}
    \label{dyn_img:zju}
\end{figure}

 
\vspace{6pt}
\noindent\textbf{Real-World Dataset.} 
We compare our method against RigGS~\cite{yao2025riggs} and AP-NeRF~\cite{uzolas2023template} on the real-world ZJU-MoCap~\cite{peng2021neural} dataset. In \tabref{dyn_table:zju}, the reported results of RigGSS~\cite{yao2025riggs} and AP-NeRF~\cite{peng2021neural} are obtained using all available time steps in the standard monocular video setup, whereas our method runs with only $1/10$ and $1/5$ of the time steps. Despite having access to significantly fewer temporal observations, our approach achieves comparable performance to these SOTA methods. We show in \figref{dyn_img:zju} that our method is able to reconstruct the motion accurately. 

\begin{table}[t!]
\centering
\caption{Results on the real-world ZJU-MoCap dataset. Note that existing methods are trained with full monocular video sequences, whereas our method uses only 10$\times$ and 5$\times$ fewer frames.}
\resizebox{0.9\columnwidth}{!}{
\begin{tabular}{cccc}
\hline
Method  & SSIM $\uparrow$ & PSNR $\uparrow$  & LPIPS ${(\times100)} \downarrow$ \\ \hline
AP-NeRF~\cite{uzolas2023template} & 0.919 & 25.62 & 9.34         \\
RigGS~\cite{yao2025riggs}   & 0.975 & 33.54 & 3.27         \\ \hline
Ours $(10\times)$   & 0.934 & 28.13 & 6.53         \\
Ours $(5\times)$   & 0.944 & 28.83 & 5.89         \\ \hline
\end{tabular}
}
\label{dyn_table:zju}
\vspace{-5pt}
\end{table}

\subsection{Relaxing the Need for Multi-View Initialization with a Pretrained Generative Model}
\label{dyn_sec:diffusion}

We demonstrate that the multi-view initialization at the canonical (static) state can potentially be replaced with a pretrained diffusion-based generative model, using only a single observation $I^r$ at the first time step.
Given $I^r$, we optimize the initial $\myset{G}$ with $\mathcal{L}_{perceptual}$ only at the corresponding viewpoint, and employ the $\mathcal{L}_{SDS}$~\cite{pooledreamfusion} to optimize all other unseen viewpoints. $\mathcal{L}_{SDS}$ is defined as:

\begin{equation}
\resizebox{.88\hsize}{!}{$
\nabla_\myset{G}\mathcal{L}_{SDS} = \mathbb{E}_{t,p,\epsilon} \left[ w(t) (\epsilon_\phi(I^p; t,I^r,\Delta p) - \epsilon)  \frac{\partial I^p}{\partial \myset{G}} \right]
$}
\end{equation}
where $w(t)$ is the weighting function from DDIM~\cite{song2020denoising} and $\epsilon_\phi(\cdot)$ is the predicted noise from a pre-trained 2D diffusion model. We use Zero-1-to-3~\cite{liu2023zero} as the diffusion prior, conditioned on $I^r$ and the relative camera pose $\Delta p$ from the reference viewpoint $r$ to the rendering viewpoint $p$. 
After the canonical $\myset{G}$ is initialized, we follow the same process described in \secref{dyn_sec:optimization} to optimize our deformation field. Since the initial $\myset{G}$ can be noisy with only one observed image, we keep $\mathcal{L}_{SDS}$ in the loss function (\eref{dyn_eq:loss_func}) during the optimization to regularize the reconstruction. More details can be found in the supplementary material.

\begin{figure}[t!]
    \centering
    \includegraphics[width=\columnwidth,trim={2cm 4cm 5.8cm 4cm},clip]{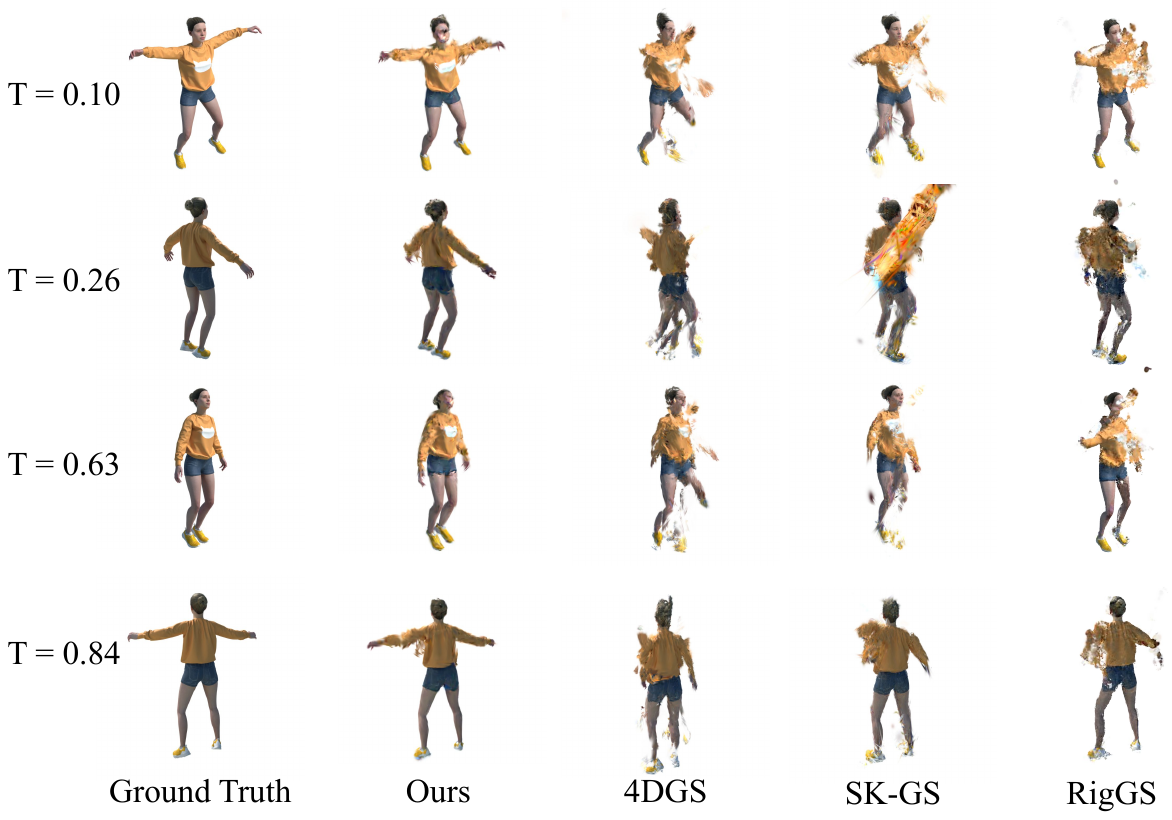}
    \caption{Comparison of all methods without access to multi-view images at the initial time step. Despite using only $11$ sparse input, our method reconstructs motion and preserves object structure more faithfully, whereas baselines are prone to artifacts under self-occlusion and sparse supervisions.}
    \label{dyn_img:jumpingjacks}
\end{figure}

We first present results on the \textit{Jumpingjacks} scene from D-NeRF~\cite{pumarola2021d} in \figref{dyn_img:jumpingjacks}. 
All methods are trained without access to the multi-view images. Despite using only $11$ input images across the entire motion sequence, our method produces more coherent and structurally consistent motion compared to the baselines. We observe that while the baselines fit the input frames well, the sparse observations and self-occlusions lead to inconsistent geometry and unrealistic deformations when viewed from unseen viewpoints.

\begin{figure}[t!]
    \centering
    \includegraphics[width=\columnwidth,trim={3.2cm 4.0cm 6.2cm 4cm},clip]{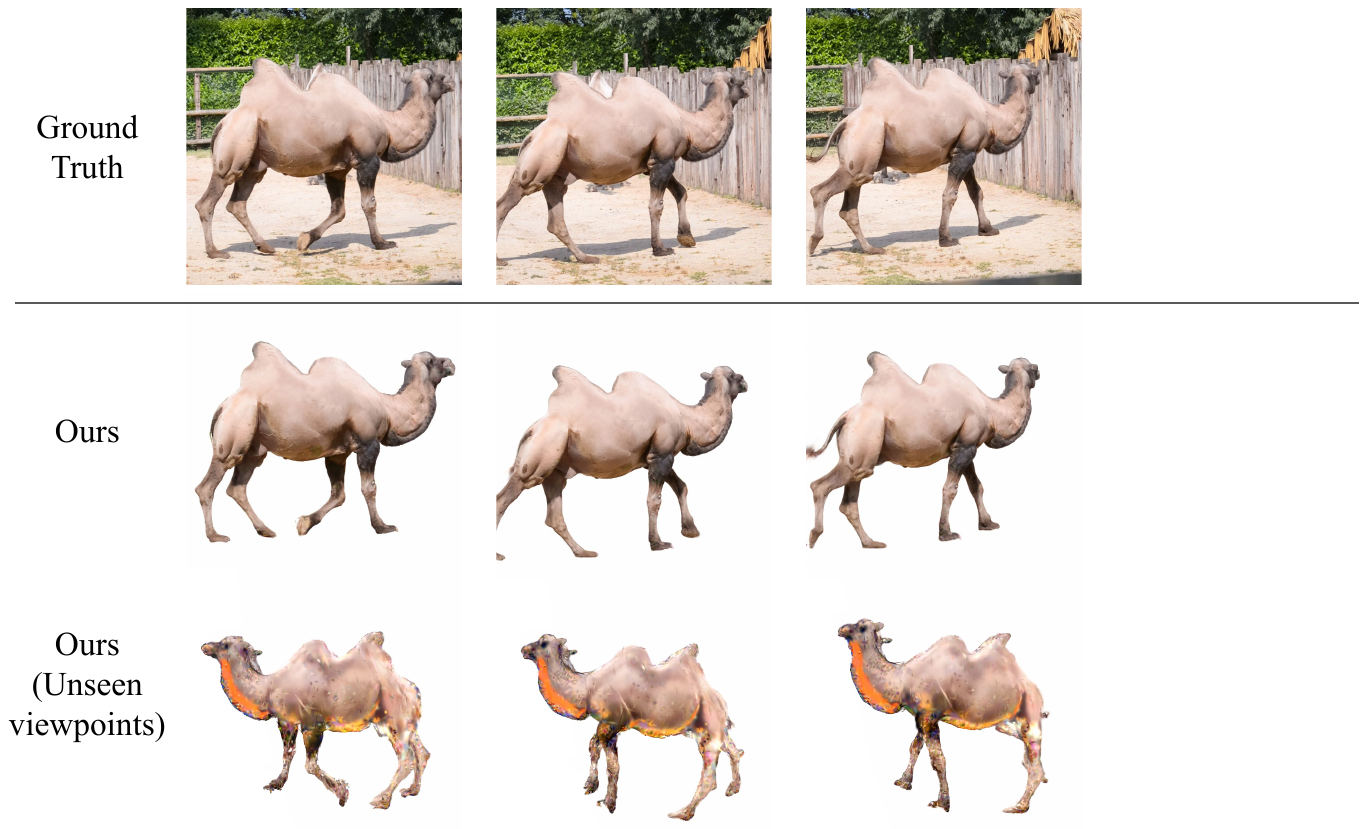}
    \caption{Results on the camel scene from the in-the-wild DAVIS dataset~\cite{Perazzi_CVPR_2016} without camera pose information. Note that this is a monocular video with fixed camera and the other side of the target is never seen in the video.}
    \label{dyn_img:camel}
    \vspace{-5pt}
\end{figure}

Additionally, we evaluate our method on the in-the-wild \textit{camel} scene from the DAVIS dataset~\cite{Perazzi_CVPR_2016}. 
Note that the other side of the target is never seen in this monocular video.
Assuming the camera is fixed across the whole sequence, we have only sparse temporal observations from a fixed viewpoint. Despite the challenging setup, we show in \figref{dyn_img:camel} that our method, paired with $\mathcal{L}_{SDS}$, successfully reconstructs plausible motion and texture for the visible regions. For the completely unseen part, the overall motion and structure are preserved, while there is oversaturated texture near the edge, which is a known issue of $\mathcal{L}_{SDS}$~\cite{mcallister2024rethinking,alldieck2024score,lukoianov2024score}.


\subsection{Ablation studies}
\label{dyn_sec:ablation}

\begin{figure}[t!]
    \centering
    \includegraphics[width=\columnwidth,trim={3.5cm 7.0cm 5.3cm 5.5cm},clip]{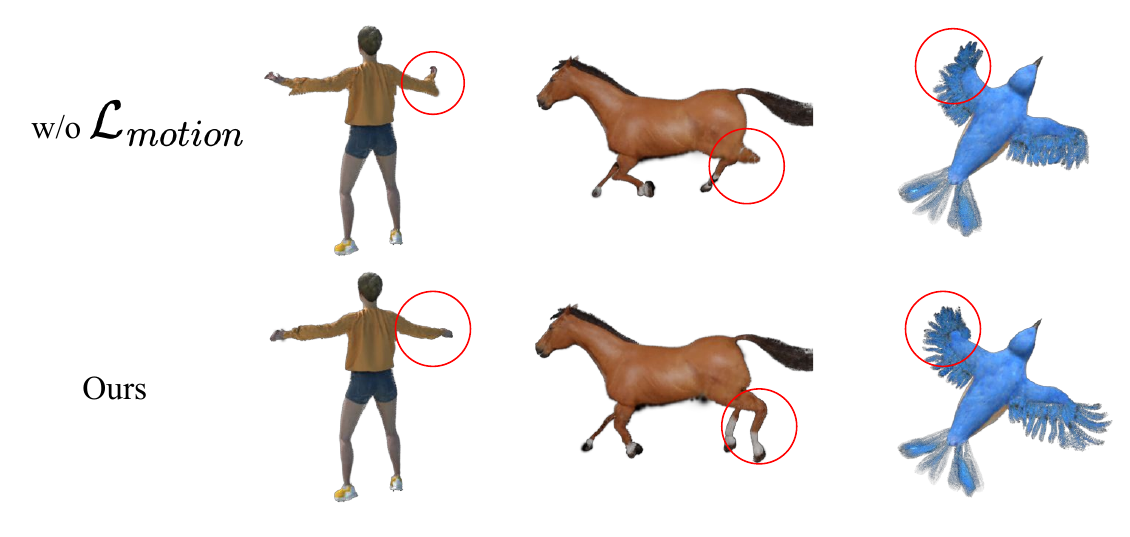}
    \caption{Qualitative comparison of results with and without $\mathcal{L}_{motion}$. This motion regularization term helps reduces noise in joint pose predictions.}
    \label{dyn_img:ablation}
    
\end{figure}

We conduct ablation studies to evaluate the effect of key components in our framework: the motion regularization term $\mathcal{L}_{motion}$, the skinning weight correction field $MLP_\Phi$, and the detail deformation field $MLP_\Psi$.
As shown in \tabref{dyn_table:ablation}, both the skinning correction field and detail deformation field contribute to improving rendering quality on the D-NeRF dataset. The skinning correction field refines the learned skinning weights when the RBF-based bone representation is insufficient, while the detail deformation field adjusts the Gaussian primitives for parts that cannot be fully explained by the learned LBS deformation. 
Although $\mathcal{L}_{motion}$ has small impact on quantitative metrics, \figref{dyn_img:ablation} shows that it reduces noise in the joint poses predicted by $MLP_\Theta$, resulting in smoother and more stable motion.

\begin{table}[t!]
\centering
\caption{Ablation study on the D-NeRF dataset. We evaluate the effect of the motion regularization term $\mathcal{L}_{motion}$, skinning correction field $MLP_\Phi$, and the detail deformation field $MLP_\Psi$.}
\resizebox{1.0\columnwidth}{!}{
\begin{tabular}{cccc}
\hline
Method  & SSIM $\uparrow$ & PSNR $\uparrow$  & LPIPS ${(\times100)} \downarrow$ \\ \hline
w/o $\mathcal{L}_{motion}$ & 0.942 & 27.26 & 6.08         \\
w/o $MLP_\Phi$  & 0.945 & 27.28 & 5.97        \\ 
w/o $MLP_\Psi$  & 0.931 & 26.34 & 6.51        \\
Ours   & \textbf{0.950} & \textbf{27.75} & \textbf{5.79}         \\ \hline
\end{tabular}
}
\label{dyn_table:ablation}
\vspace{-5pt}
\end{table}

%% file: dyn_sec/conclusion.tex
\section{Conclusion and Future Work}
\label{dyn_sec:conclusion}

We presented \ourmethod, a method for articulated dynamic object reconstruction from sparse temporal observations. \ourmethod{} leverages a rough input skeleton and an initial static reconstruction to learn a skeleton-driven deformation field that models coherent motion across time.
Furthermore, we showed that the need for multi-view initialization can be relaxed using a pre-trained diffusion-based generative prior, enabling dynamic reconstruction in real-world scenarios.
Experiments on synthetic datasets show that \ourmethod{} outperforms existing methods by up to 34\% in PSNR under sparse observations and performs comparably to dense monocular methods on real-world datasets, even though \ourmethod{} uses $10\times$ fewer frames. 
While promising, our approach has limitations. The diffusion-based initialization can fail under severe self-occlusion or uncommon viewpoints, as it relies on a general pre-trained model. 
Moreover, test-time interpolation may struggle with highly complex motion. 
A potential future direction is to investigate using  category-specific priors or a pre-trained prior conditioned on the noisy skeleton input to guide motion estimation and reconstruction.